%% file: le2ed.tex
\documentclass[letterpaper,10pt,conference]{ieeeconf}

\IEEEoverridecommandlockouts                            
\input{packages}
\input{macros}

\title{\LARGE \bf
Efficient and Robust LiDAR-Based End-to-End Navigation
}

\author{Zhijian Liu$^{*,1}$, Alexander Amini$^{*,2}$, Sibo Zhu$^{1}$, Sertac Karaman$^{3}$, Song Han$^{1}$, and Daniela L. Rus$^{2}$
\thanks{$*$ The first two authors have contributed equally to this work with order determined by a coin toss. This work was supported by National Science Foundation and Toyota Research Institute. We gratefully acknowledge the support of NVIDIA with the donation of V100 GPU and Drive AGX Pegasus.}%
\thanks{$^{1}$ Microsystems Technology Laboratories (MTL), Massachusetts Institute of Technology {\tt\{zhijian,sibozhu,songhan\}@mit.edu}}%
\thanks{$^{2}$ Computer Science and Artificial Intelligence Lab (CSAIL), Massachusetts Institute of Technology {\tt\{amini,rus\}@mit.edu}}%
\thanks{$^{3}$ Laboratory for Information and Decision Systems (LIDS), Massachusetts Institute of Technology {\tt\{sertac\}@mit.edu}}%
}

\begin{document}

\maketitle
\thispagestyle{empty}
\pagestyle{empty}

\input{text/abstract}
\input{text/intro}
\input{text/related}
\input{text/method}
\input{text/setup}
\input{text/results}
\input{text/conclusion}

\bibliographystyle{IEEEtran}
\bibliography{reference}

\end{document}

%% file: packages.tex
\usepackage{adjustbox}
\usepackage{array}
\usepackage{booktabs}
\usepackage{colortbl}
\usepackage{float,wrapfig}
\usepackage{hhline}
\usepackage{multirow}
\usepackage{subcaption}
\usepackage{cite}

\usepackage{amsmath,amsfonts,amssymb}
\usepackage{bm}
\usepackage{nicefrac}
\usepackage{microtype}
\usepackage{inconsolata}

\usepackage{changepage}
\usepackage{extramarks}
\usepackage{fancyhdr}
\usepackage{lastpage}
\usepackage{setspace}
\usepackage{soul}
\usepackage{xspace}
\usepackage{paralist}

\usepackage{url}

\usepackage{algorithm,algpseudocode}
\usepackage{enumerate}
\usepackage{lipsum}

\usepackage{titlesec}
\usepackage{listings}

\usepackage{dblfloatfix}

%% file: macros.tex
\newcolumntype{L}[1]{>{\raggedright\let\newline\\\arraybackslash\hspace{0pt}}m{#1}}
\newcolumntype{C}[1]{>{\centering\let\newline\\\arraybackslash\hspace{0pt}}m{#1}}
\newcolumntype{R}[1]{>{\raggedleft\let\newline\\\arraybackslash\hspace{0pt}}m{#1}}

\newcommand{\sect}[1]{Section~\ref{#1}}

\newcommand{\eqn}[1]{Equation~\ref{#1}}

\newcommand{\fig}[1]{Figure~\ref{#1}}

\newcommand{\tab}[1]{Table~\ref{#1}}

\newcommand{\ignorethis}[1]{}
\newcommand{\norm}[1]{\lVert#1\rVert}

\makeatletter
\DeclareRobustCommand\onedot{\futurelet\@let@token\@onedot}
\def\@onedot{\ifx\@let@token.\else.\null\fi\xspace}

\def\eg{\emph{e.g}\onedot} 
\def\ie{\emph{i.e}\onedot}

\def\etal{\emph{et al}\onedot}
\makeatother

\definecolor{citecolor}{RGB}{34,139,34}
\definecolor{mydarkblue}{rgb}{0,0.08,1}
\definecolor{mydarkgreen}{rgb}{0.02,0.6,0.02}
\definecolor{mydarkred}{rgb}{0.8,0.02,0.02}
\definecolor{mydarkorange}{rgb}{0.40,0.2,0.02}
\definecolor{mypurple}{RGB}{111,0,255}
\definecolor{myred}{rgb}{1.0,0.0,0.0}
\definecolor{mygold}{rgb}{0.75,0.6,0.12}
\definecolor{mydarkgray}{rgb}{0.66, 0.66, 0.66}

\def\model{Fast-LiDARNet\xspace}
\def\fusion{Hybrid Evidential Fusion\xspace}

\algnewcommand\algorithmicswitch{\textbf{switch}}
\algnewcommand\algorithmiccase{\textbf{case}}
\algnewcommand\algorithmicassert{\texttt{assert}}
\algnewcommand\Assert[1]{\State \algorithmicassert(#1)}%

\algdef{SE}[SWITCH]{Switch}{EndSwitch}[1]{\algorithmicswitch\ #1\ \algorithmicdo}{\algorithmicend\ \algorithmicswitch}%
\algdef{SE}[CASE]{Case}{EndCase}[1]{\algorithmiccase\ #1}{\algorithmicend\ \algorithmiccase}%
\algtext*{EndSwitch}%
\algtext*{EndCase}%

%% file: text/abstract.tex
\begin{abstract}

Deep learning has been used to demonstrate end-to-end neural network learning for autonomous vehicle control from raw sensory input. 
While LiDAR sensors provide reliably accurate information, existing end-to-end driving solutions are mainly based on cameras since processing 3D data requires a large memory footprint and computation cost.
On the other hand, increasing the robustness of these systems is also critical; however, even estimating the model's uncertainty is very challenging due to the cost of sampling-based methods. 
In this paper, we present an efficient and robust LiDAR-based end-to-end navigation framework. 
We first introduce \model that is based on sparse convolution kernel optimization and hardware-aware model design. 
We then propose \fusion that directly estimates the uncertainty of the prediction from only a single forward pass and then fuses the control predictions intelligently.
We evaluate our system on a full-scale vehicle and demonstrate lane-stable as well as navigation capabilities. 
In the presence of out-of-distribution events (\eg, sensor failures), our system significantly improves robustness and reduces the number of takeovers in the real world.

\end{abstract}

%% file: text/intro.tex
\section{Introduction}
\label{sect:intro}

End-to-end learning has produced promising solutions for reactive or instantaneous control of autonomous vehicles directly from raw sensory data~\cite{bojarski2016end,amini2018variational,codevilla2018end,lechner2020neural}. Furthermore, recent works have demonstrated the ability to perform point-to-point navigation using only coarse localization and sparse topometric maps~\cite{amini2019variational,hawke2020urban}, without the need for pre-collected high-definition maps of an environment. While these approaches rely heavily on vision data from cameras, LiDAR sensors could provide more accurate distance (depth) information and greater robustness to environmental changes like illumination. 

However, learning 3D representations from LiDAR efficiently for end-to-end control networks remains challenging, due to the unordered structure and the large size of LiDAR data: \eg, a 64-channel LiDAR sensor produces more than 2 million points per second. As state-of-the-art 3D networks~\cite{choy20194d} require 14$\times$ more computation at inference time than their 2D image counterparts~\cite{he2016deep}, many researchers have explored 2D-based solutions that operate on spherical projections~\cite{cortinhal2020salsanext} or bird's-eye views~\cite{zhang2020polarnet} to save computation and memory. Though efficient, reduction to 2D introduces a critical loss of geometric information. Recent 3D neural networks~\cite{choy20194d,tang2020searching,liu2020hardware} process 3D input data using sparse convolutions~\cite{graham20183d} to avoid the problem of cubically-growing memory consumption. However, these methods are not favored by modern high-parallelization hardware due to the irregular memory access pattern introduced by sparse convolution~\cite{liu2019point}. To achieve safe and efficient LiDAR-based end-to-end driving, novel, hardware-aware 3D neural architectures are required. 

\input{figText/teaser}

Real-world deployable robotic systems must not only be accurate and efficient, but also highly robust. End-to-end models typically predict instantaneous control and generally suffer from the high sensitivity to perturbations (\eg, noisy sensory inputs). To address this, a plausible solution is to integrate several consecutive frames as input~\cite{xu2017end}. However, this is neither efficient in the 3D domain nor effective in closed-loop control settings, since it requires either modeling recurrence~\cite{hochreiter1997long} or applying 4D convolutions~\cite{choy20194d} to process multiple input frames. Instead, by training a model to additionally predict \textit{future} control and apply odometry-corrected fusion, actuation commands could potentially be stabilized. Furthermore, capturing the model's confidence associated with each prediction could enable even greater robustness by accounting for potential ambiguities in future behavior as well as unexpected out-of-distribution (OOD) events (\eg, sensor failures).

In this work, we present a novel LiDAR-based end-to-end navigation system capable of achieving efficient and robust control of a full-scale autonomous vehicle using only raw 3D point clouds and coarse-grained GPS maps. Specifically, we present the following contributions:
\begin{enumerate}
    \item \emph{\model}, an efficient, hardware-aware, and accurate neural network that operates on raw 3D point clouds with accelerated sparse convolution kernels and runs at 11 FPS on NVIDIA Jetson AGX Xavier;
    \item \emph{\fusion}, a novel uncertainty-aware fusion algorithm that directly learns prediction uncertainties and adaptively integrates predictions from neighboring frames to achieve robust autonomous control; 
    \item Deployment of our system on a full-scale autonomous vehicle and demonstration of navigation and improved robustness in the presence of sensor failures.
\end{enumerate}

%% file: figText/teaser.tex
\begin{figure}[t!]
\centering
\includegraphics[width=\linewidth]{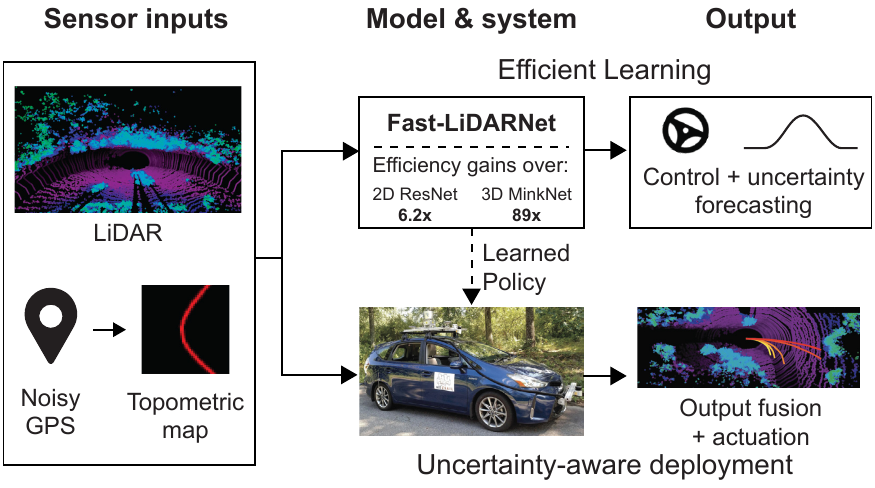}
\caption{\textbf{End-to-end navigation with LiDAR}. An end-to-end \model is efficiently designed, using only LiDAR and topometric map inputs, to predict a control output and uncertainties with low computation. Here, efficiency gains are measured by the FLOPs reduction. During deployment, a novel fusion algorithm incorporates future predictions and utilizes uncertainties for robust steering actuation.}
\label{fig:teaser:overview}
\vspace{-15pt}
\end{figure}

%% file: text/related.tex
\section{Related Work}
\label{sect:related}

\input{figText/overview}

\textbf{End-to-End Autonomous Driving. }
Map-based navigation of autonomous vehicles relies heavily on pre-collected high-definition (HD) maps using either LiDAR~\cite{burgard2008map, levinson2010robust} and/or vision~\cite{wolcott2014visual} to precisely localize the robot~\cite{leonard1991simultaneous, montemerlo2002fastslam}. These HD maps are expensive to create, large to maintain, and difficult to process efficiently. Recent works on ``map-lite'' approaches~\cite{ort2018autonomous, ort2019maplite} require only sparse topometric maps which are orders of magnitude smaller and openly available~\cite{haklay2008openstreetmap}. However, these approaches are largely rule-based and do not leverage modern advances in end-to-end learning of control representations. Our work builds on the advances of end-to-end learning of reactionary control~\cite{bojarski2016end, xu2017end} and navigation~\cite{amini2019variational, codevilla2018end, hawke2020urban} and extends beyond 2D sensing. Recent works have investigated multi-modal fusion of vision and depth to improve control~\cite{xiao2020multimodal, patel2017sensor, bohez2017sensor}. However, these approaches project 3D data onto 2D depth maps to leverage 2D convolutions, thereby losing significant geometric information. We hypothesize that LiDAR-based end-to-end networks can learn subtle details in scene geometry, without the need for HD maps, and achieve robust deployment in the real world. 

\textbf{Hardware-Efficient LiDAR Processing. }
Real-world robotic systems, such as autonomous vehicles, often face significant computational resource challenges. LiDAR processing and 3D point cloud networks~\cite{qi2017pointnet, wang2019dynamic, choy20194d} make this extremely challenging due to their increased computational demand. Efficiency can be improved by aggressively downsampling the point cloud~\cite{hu2020rand}, leveraging more efficient memory representations in less dense regions~\cite{riegler2017octnet, wang2017cnn, wang2018adaptive, lei2019octree}, or using sparse convolutions to accelerate vanilla volumetric operations~\cite{graham20183d, choy20194d}. However, all these methods still require a large amount of random memory accesses, which are difficult to parallelize and very inefficient on modern hardware. We propose a novel and efficient method capable of learning from full LiDAR point clouds while successfully deployed on a full-scale vehicle, running at more than 10 FPS.

\textbf{Uncertainty-Aware Learning. }
Robots must be able to deal with high amounts of uncertainty in their environment~\cite{djuric2020uncertainty}, sensory data~\cite{gilitschenski2019deep}, and decision making~\cite{michelmore2018evaluating}. This involves estimating predictive uncertainties and systematically leveraging uncertainties together with control decisions to increase robustness. Bayesian deep learning~\cite{kendall2018learning, gal2016dropout, lakshminarayanan2017simple, hernandez2015probabilistic} has emerged as a compelling approach to estimate epistemic (\ie, model) uncertainty by placing priors over network weights and stochastically sampling to approximate uncertainty. However, sampling approaches are infeasible in resource constrained environments, because in robotics, due to their computational expense. Instead, evidential deep learning aims to directly learn the underlying epistemic uncertainty distribution using a neural network to estimate uncertainty without the need for sampling~\cite{amini2020deep, sensoy2018evidential}. We demonstrate how evidential learning methods can be integrated into an end-to-end learning system to execute uncertainty-aware control.

%% file: figText/overview.tex
\begin{figure*}[!t]
\centering
\includegraphics[width=\linewidth]{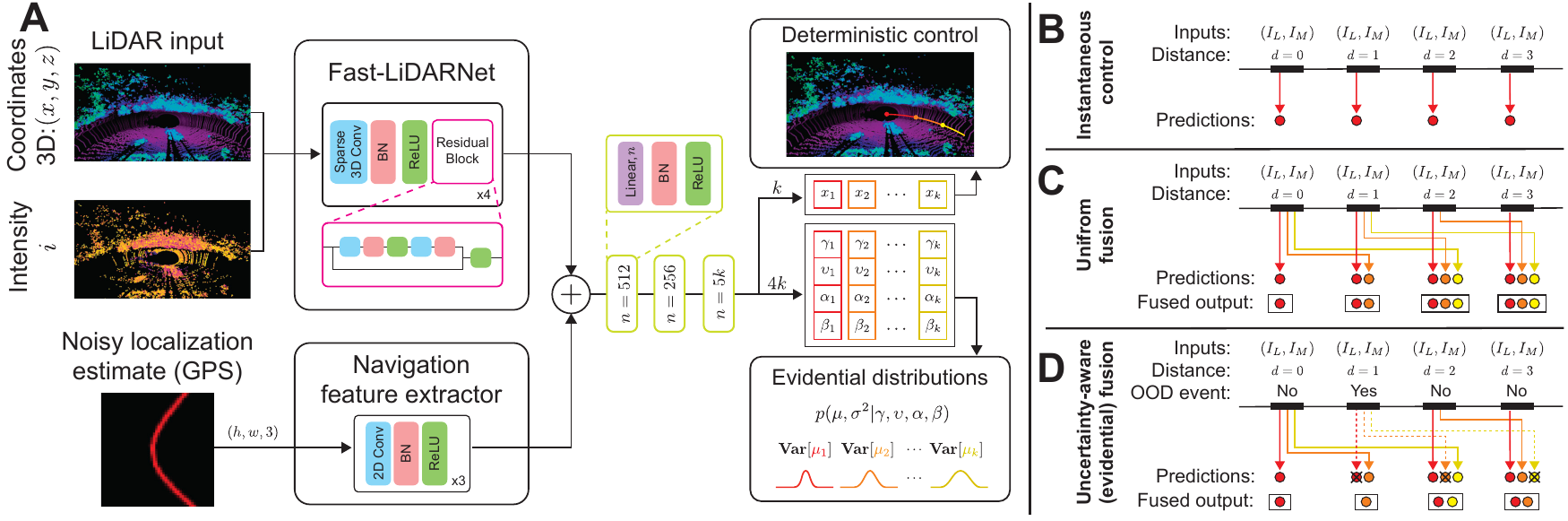}
\caption{\textbf{Overview of our solution.} \textbf{(A)} Raw LiDAR point clouds (the visualized colors are based on heights and intensities) and noisy roadmaps are fed to \model and navigation feature extractor, and integrated to learn both a deterministic control value as well as the parameters of a higher-order evidential distribution capturing the underlying predictive distribution. Output control predictions can be (\textbf{B}) instantaneously executed or (\textbf{C}) uniformly fused with odometry-corrected past predictions. (\textbf{D}) Uncertainty estimation using our evidential outputs enables intelligently weighting our predictions to increase the robustness, especially on out-of-distribution (OOD) events, or on increased uncertainty of ambiguous future time steps.}
\label{fig:overview}
\vspace{-15pt}
\end{figure*}

%% file: text/method.tex
\section{Methodology}
\label{sect:method}

Traditional end-to-end models perform ``reactive'' control for instantaneous execution. By training the model to additionally predict control commands at steps into the future, we can ensemble and fuse these predictions once the robot reaches those points, thereby stabilizing control. This paper takes this idea even further and intelligently fuses the control predictions according to the model's uncertainty at each point, in order to deal with sudden unexpected events or an ambiguous future environment (\ie, highly uncertain events). 

Given a raw LiDAR point cloud $I_\text{L}$, and a rendered bird's-eye view image of the noisy, routed roadmap $I_\text{M}$, our objective is to learn an end-to-end neural network $f_\theta$ to directly predict the control signals that can drive the vehicle as well as the corresponding epistemic (model) uncertainty.
\begin{equation}
    \{(x_k, \bm{e}_k)\}_{k=0}^{K-1} = f_\theta(I_\text{L}, I_\text{M}),
\end{equation}
where the network outputs $K$ predictions, each pair of $x_k$ and $\bm{e}_k$ corresponding to a control predictions with future lookahead distance of $k$ [m] from the current frame: $x_k$ is the predicted control value (which can be supervised by the recorded human control $y_k$), and $\bm{e_k}$ are the hyperparameters to estimate the uncertainty of this prediction. In principle, we can depend on the current control $x_0$ alone to drive the vehicle. However, the remaining $x_k$'s with $k > 0$ might also be used to improve the robustness of the model with uncertainty-weighted temporal fusion (using $\bm{e}_k$'s); learning these additional $x_k$'s also provides the model with a sense of planning and predicting the future.

We illustrate an efficient and robust LiDAR-only end-to-end neural network in \fig{fig:overview}. The following subsections describe our training and deployment methodology in two parts. First, we describe how we learn efficient representations directly from input point cloud data $I_\text{L}$ (\sect{sect:method:lidar}) and a rough routed map $I_\text{M}$ (following \cite{amini2019variational}). These perception (LiDAR) and localization (map) features are combined and fed to a fully-connected network to predict our control and uncertainty estimates. Next, in \sect{sect:method:fusion}, we describe how our network can be optimized to learn its uncertainty and present a novel algorithm for leveraging this uncertainty during deployment to increase the robustness of the robot. 

\subsection{\model: Efficient LiDAR Processing}
\label{sect:method:lidar}

Most recent 3D neural networks~\cite{choy20194d} apply a stack of sparse convolutions to process the LiDAR point cloud. These sparse convolutions only store and perform computation at non-zero positions to save the memory consumption and computation. However, they are not favored by the modern hardware. We propose several optimizations to accelerate the sparse CUDA kernel. (1) The index lookup of sparse convolution (to query the index given the coordinate) is usually implemented with a multi-thread CPU hashmap~\cite{choy20194d}. This is bottlenecked by the limited parallelism of CPUs. To solve this issue, we map the hashmap-based index lookup implementation onto GPUs with Cuckoo hashing~\cite{pagh2001cuckoo} as it parallelizes well for both construction and query. This optimization reduces the overhead of kernel map construction to only 10\%. (2) The sparse computation suffers from the high cost of irregular memory access as adjacent points in the 3D space may not be adjacent in the sparse tensor. To tackle this, we coalesce the memory access before convolution via a gathering procedure (similar to \texttt{im2col}) and perform the matrix multiplication (\texttt{GEMM}) on a large, regular matrix. 

Apart from sparse kernel optimization, we further redesign the LiDAR processing network for more acceleration. For 2D CNNs, channel pruning is a common technique to reduce the latency~\cite{he2017channel,he2018amc}. However, as the sparse convolution is bottlenecked on data movement, not matrix multiplication, reducing the channel numbers is less effective: when reducing the channel numbers in MinkowskiNet~\cite{choy20194d} by half, the latency is only reduced by 1.1$\times$ (although in theory the reduction is 4$\times$). This is due to computation reduces quadratically with the pruning ratio, while memory only reduces linearly. Realizing this bottleneck, we jointly reduce the input resolution, channel numbers and network depths, rather than just pruning the channel dimension. We choose these two dimensions because both computation and memory access cost are linear to the number of input points (resolution) and layers (depth). As a result, \model operates on the input with voxel size of 0.2m and is composed of four residual sparse convolution blocks (with 16, 16, 32 and 64 channels, respectively) together. Within each block, there are two 3D sparse convolutions to extract the local features. We also apply an additional sparse convolution with stride of 2 after each block to downsample the point cloud to enlarge the perceptive field. 

\subsection{Hybrid Evidential Fusion}
\label{sect:method:fusion}

\fig{fig:overview} illustrates that one of our model's output branches directly predicts the control values $\{x_k\}$, which can simply be supervised with the L1 loss: $\mathcal{L}_\text{MAE}(x_k, y_k) = \norm{x_k - y_k}_1$. During deployment, we use the predictions from both current and odometry-corrected previous frames to improve stability. For each frame, we keep track of its absolute travelled distance $d$ and denote its corresponding predictions as $\{x^{(d)}_i\}_{i=0}^{K-1}$. As the accuracy of $d$ is important only locally (within the $K$ lookahead distances), we estimate $d$ from odometry based on an Extended Kalman Filter~\cite{julier2004unscented}. We can then fuse
\begin{equation}
    \mathcal{X}^{(d)} = \left\{ x^{(d)}_0, x^{(d-1)}_1, x^{(d-2)}_2, \ldots, x^{(d-i)}_i, \ldots \right\}
\end{equation}
to obtain the estimated control for the current frame, as they are all estimates of this frame. One straightforward way of fusion is to directly average all predictions in $\mathcal{X}^{(d)}$, including the current instantaneous prediction, as well as previous future predictions. However, this approach neglects the uncertainty of the predictions. This is important because (1) predictions from the past are usually less accurate due to ambiguity; and (2) out-of-distribution (OOD) events (\eg, sudden changes, sensor failures) can make the current prediction extremely important or completely wrong.

To address this issue, our model additionally learns to estimate the epistemic uncertainty of every prediction, rather than relying on the rule-based fusion. This is accomplished by training an additional branch of the network to output an evidential distribution~\cite{amini2020deep,sensoy2018evidential} for each prediction. Evidential distributions aim to capture the evidence (or confidence) associated to any decision that a network makes, thus allowing us to intelligently weigh less-confident predictions by the network less than those which are more confident. We assume that our control labels, $y_d$, are drawn from an underlying Gaussian Distribution with unknown mean and variance $(\mu, \sigma^2)$, which we seek to probabilistically estimate. As for regression targets, like robotic control problems, we place priors over the likelihood variables to obtain a joint distribution, $p(\mu, \sigma^2 | \gamma, \upsilon, \alpha, \beta)$ with
\begin{equation}
\mu \sim \mathcal{N}(\gamma, \sigma^2 \upsilon^{-1}),  \quad\qquad \sigma^2 \sim \Gamma^{-1}(\alpha, \beta).
\end{equation}
Our network is trained to output the hyperparameters defining this distribution, $\bm{e}_k = (\gamma_k, \upsilon_k, \alpha_k, \beta_k)$, by jointly maximizing model fit ($\mathcal{L}_\text{NLL}$) and minimizing evidence on errors ($\mathcal{L}_\text{R}$):
\begin{equation}
\begin{aligned}
    &\mathcal{L}_\text{NLL}(\bm{w}_k, y_k) = \tfrac{1}{2} \log\left(\tfrac{\pi}{\nu_k}\right) - \alpha_k\log(\Omega_k) \\
    &+ \left( \alpha_k + \tfrac{1}{2} \right) \log\left( (y_k - \gamma_k)^2\nu_k + \Omega_k \right) + \log\left(\tfrac{\Gamma(\alpha_k)}{\Gamma(\alpha_k + 1/2)}\right) \\
    &\mathcal{L}_\text{R}(\bm{w}_k, y_k) = |y_k - \gamma_k| \cdot (2\alpha_k + \nu_k)
\end{aligned}
\end{equation}
where $\Omega_k = 2\beta_k(1 + \nu_k)$. We refer the readers to Amini~\etal~\cite{amini2020deep} for more details about evidential regression. Finally, all losses are summed together to compute the total loss:
\begin{equation}
\begin{aligned}
\label{eqn:loss}
    \mathcal{L}(\cdot) = \sum\nolimits_k \big(\alpha\mathcal{L}_\text{MAE}(\cdot) + \mathcal{L}_\text{NLL}(\cdot) + \mathcal{L}_\text{R}(\cdot)\big)
\end{aligned}
\end{equation}
where $\alpha = 1000$ for weighting scale differences. After optimization, the epistemic uncertainty can be directly computed as $\text{Var}[x_k] = \beta_k / (\nu_k(\alpha_k - 1))$ without sampling~\cite{amini2020deep}. During deployment, we can collect the deterministic prediction $x_k$ as well as the corresponding evidential uncertainty $\text{Var}[x_k]$ from previous frames. We can then use the uncertainty to evaluate a confidence-weighted average of our predictions:

\begin{figure}[h]
\vspace{-15pt}
\centering
\begin{minipage}[!h]{\linewidth}
\begin{algorithm}[H]
\setstretch{1.15}
\caption{Uncertainty-aware deployment}
\begin{algorithmic}
    \small
    \State \textbf{Given}: policy $f_\theta$, inputs ($I_\text{L}, I_\text{M}$), and fusion method 
    \State $\left\{(x_k, \gamma_k, \upsilon_k, \alpha_k, \beta_k)\right\} \leftarrow f_\theta(I_\text{L}, I_\text{M})$ \Comment{Inference}
    \For {$k \in \{0, 1, \dotsc, K-1\}$}
        \State Var$[\mu_k] \leftarrow \beta_k / (\upsilon_k (\alpha_k-1))$ \Comment{Compute uncertainty}
        \State $\lambda_k \leftarrow 1 \,/\, \text{Var}[\mu_k] $ \Comment{Compute confidence}
        \State $\Lambda^{(d+k)} \leftarrow \Lambda^{(d+k)} \cup\, \{\lambda_k\}$ \Comment{Store confidence}
        \State $\mathcal{X}^{(d+k)} \leftarrow \mathcal{X}^{(d+k)} \cup\, \{x_k\}$ \Comment{Store prediction}
    \EndFor
    \State $\Lambda^{(d)} \leftarrow \Lambda^{(d)} / \sum_{\lambda\in\Lambda^{(d)}} \lambda$ \Comment{Normalize confidence}
    \Switch{fusion}
        \Case{none:} \Comment{Instantaneous (no fusion)}
            \State \textbf{return} $x_0^{(d)}$ 
        \EndCase
        \Case{uniform:} \Comment{Uniform fusion}
            \State \textbf{return} $(\sum_{j} \mathcal{X}^{(d)}_j) / \norm{\mathcal{X}^{(d)}}$ 
        \EndCase
        \Case{evidential:} \Comment{Evidential fusion}
            \State \textbf{return} $(\sum_{j} \mathcal{X}^{(d)}_j \Lambda^{(d)}_j) / \norm{\mathcal{X}^{(d)}}$ 
        \EndCase
    \EndSwitch
\end{algorithmic}
\label{alg:deployment}
\end{algorithm}
\end{minipage}
\vspace{-10pt}
\end{figure}

%% file: text/setup.tex
\input{figText/data}
\input{figText/results/ablation}

\section{Experimental Setup}
\label{sect:setup}

\subsection{System Setup and Data Collection}
\label{sect:system}

Our vehicle platform is a 2015 Toyota Prius V which we outfitted with autonomous drive-by-wire capabilities~\cite{naser2017parallel}. Its primary perception sensor is a Velodyne HDL-64E LiDAR, operating at 10Hz. We capture the coarse-grained global localization ($\pm 30$m) using an OxTS RT3000 GPS, and we use a Xsens MTi 100-series IMU to compute the curvature of the vehicle’s path for supervision. All vehicle computation is done onboard an NVIDIA AGX Pegasus, which is the latest generation in-car supercomputer for autonomous driving. For navigation, we gather information from OpenStreetMaps (OSM) throughout the traversed path. We render the routed bird's-eye navigation map by drawing all roads white on a black canvas, and overlay red on the traversed roads (see \fig{fig:data}). With this system set up, we collect a small-scale real-world dataset with 32km of driving (29km for training and 3km for testing) taken in a suburban area.

\subsection{Data Processing and Augmentation}
\label{sect:data-augmentation}

After the data collection, we clean and prepare the dataset for training. We first remove all frames with relatively low velocities (less than 1m/s). Following Amini~\etal~\cite{amini2019variational}, we then annotate each remaining frame with its corresponding maneuver (\ie, lane-stable, turn, junk). We use both ``lane-stable'' and ``turn'' frames to train models with navigation, and only ``lane-stable'' frames for LiDAR-only models (since there is no information of which direction to go for intersections). For each point cloud, we first filter out all points within 3m from the sensor and then quantize its coordinates with voxel size of 0.2m to reduce the number of points to around 40k.

Given the yaw rate $\gamma$ (rad/sec) and the velocity $v$ (m/sec) of the vehicle, the curvature (or inverse steering radius) of the path can be calculated as $y = \gamma / v$. For each frame, we then use the linear interpolation to obtain future curvatures $y_k$ with different lookaheads $k$. These $y_k$'s serve as supervision signals for our models. As we can easily compute the control from the curvature (with the slip angle from IMU)~\cite{amini2019variational}, we will use these two terms interchangeably throughout this paper.

We apply a number of data augmentations during training to enable the model to see more novel cases that are not collected (in \fig{fig:data}). Specifically, we first randomly scale the point cloud by a factor uniformly sampled from $[0.95, 1.05]$. We then randomly rotate the scaled point cloud by a small yaw angle $\theta \in \pm 10^\circ$. We also apply a correction to the supervised control values $\{y_k\}$ to teach the model to recover from these off-center rotations by returning to the centerline within 20 meters. For the navigation, we perturb the map with random translation ($\mathcal{N}(0, 3)$) and rotation ($\mathcal{N}(0, \pi/20)$). Besides, we also black out the map and remove the route from the map, each with probability 0.25. Note that this last augmentation is only applied to the ``lane-stable'' frames since we need to provide sufficient navigational information during the ``turn'' portion to remove ambiguity.

\subsection{Model Training}

Frames with smaller absolute control values contribute less to the optimization. This could be catastrophic as the vehicle will be off the track if the model predicts inaccurate controls in these frames. To compensate this, we multiply each term in \eqn{eqn:loss} with a scale factor $(1 + \text{exp}[-y_k^2/(2\sigma^2)])$ so that frames with smaller absolute control values will have larger loss magnitudes. We set $\sigma$ to $1/15$ in all our experiments. Optimization is carried out using ADAM~\cite{kingma2014adam} with $\beta_1 = 0.9$, $\beta_2 = 0.999$ and a weight decay of $10^{-4}$. We train all models for 250 epochs with a mini-batch size of 64. The learning rate is initialized as $3 \times 10^{-3}$ with a cosine decay schedule~\cite{loshchilov2016sgdr}.

%% file: figText/data.tex
\begin{figure}[b]
\centering
\includegraphics[width=\linewidth]{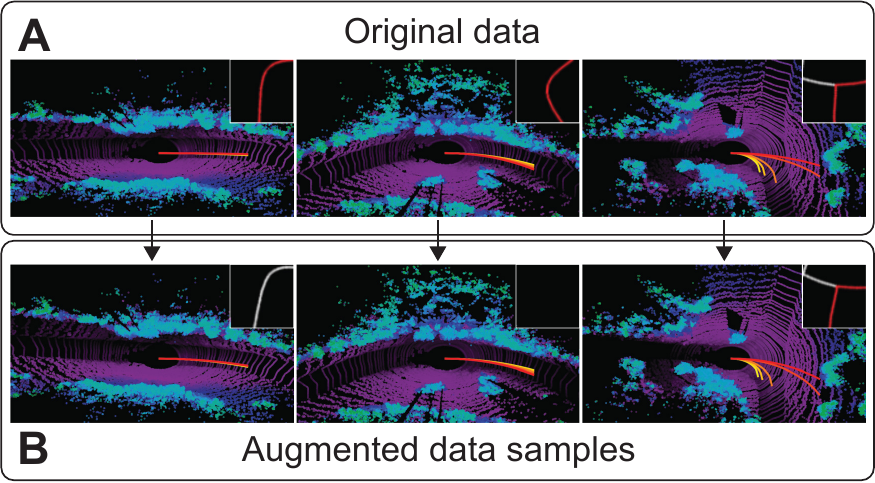}
\caption{\textbf{Improving the data efficiency by augmentation.} During training, the point clouds are randomly rotated (\eg, clockwise, left), and the supervised controls are corrected accordingly. Navigation maps are also augmented (\eg, the route can be removed, left; or the entire map dropped, middle).}
\label{fig:data}
\end{figure}

%% file: figText/results/ablation.tex
\begin{figure*}[!t]
\centering
\includegraphics[width=\linewidth]{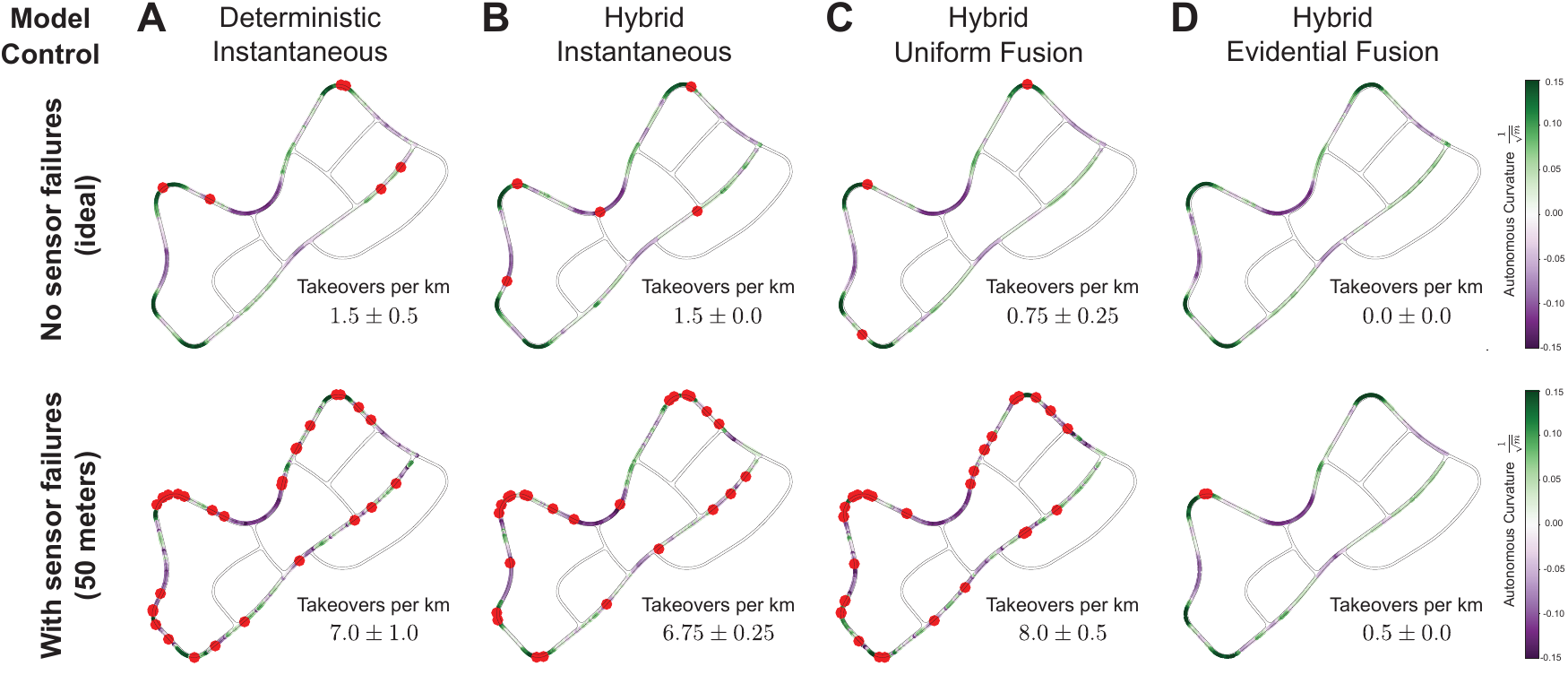}
\caption{\textbf{Real-world evaluation of LiDAR-only models.} Performance of learning models (deterministic or hybrid) and fusion strategies (instantaneous, uniform, or evidential) in scenarios without (top) or with (bottom) sensor failures. Interventions are marked by red circles, and vehicle path colored predicted curvature. Trials repeated twice at fixed speeds on the test track.}
\label{fig:results:ablation}
\vspace{-10pt}
\end{figure*}

%% file: text/results.tex
\section{Results}
\label{sect:results}

\subsection{Uncertainty-Aware Deployment of LiDAR-only Models}

We first evaluate the performance of LiDAR-only models on the lane-stable task. To test our system's robustness to out-of-distribution (OOD) events, we also manually trigger sensor failures of LiDAR every 50 meters. In \fig{fig:results:ablation}, the model deployed with (\textbf{D}) evidential fusion yields drastically improved performance with and without OOD sensor failures. This model is the most effective at not only reliably estimating high uncertainty on the OOD events, but insuring that the associated outputs will not be propagated downstream to the controller after the fusion. On the other hand, the model with (\textbf{C}) uniform fusion also obtains increased performance over instantaneous models (\textbf{A}, \textbf{B}) without sensor failures but in the presence of failures also propagates the failed response to the controller and amplifies the error by fusing it over multiple time points. Here, deterministic (\textbf{A}) and hybrid (\textbf{B}) models are compared as an ablation analysis to verify roughly similar performance regardless of output parameterization if not considering uncertainty.

\input{figText/results/augmentation}

All models in \fig{fig:results:ablation} are trained with data augmentations strategies described in \sect{sect:data-augmentation}. We also test to disable the rotation augmentation in \fig{fig:results:augmentation} to see its effect. Without rotation augmentation, the model is not able to control the vehicle to recover from off-center views because they are not encountered during data collection.

\input{figText/results/recovery}

To test the recovery robustness, we start the controller in manually distorted orientations on the road and measure the network's ability to recover (in \tab{tab:recovery}). A successful recovery is marked if the vehicle returns to a stable position within 10 seconds. We noticed that both evidence fusion and uniform fusion outperform others. All models have a better performance recovering from CCW than CW since LiDAR is not occluded by the far side of the road on a CCW turn, while the near side is occluded in a CW turn.

\input{figText/results/intersection}
\input{figText/results/intersections}

\subsection{Navigation Performance}

We evaluate the performance of navigation-enabled models on the intersections. From \fig{fig:results:navigation}A, our model performs well especially when evidential fusion is enabled, matching results from the LiDAR-only testing. It follows the direction given by the navigation map, and its trajectory is aligned closely to the human driver. We also test ablations of the map and route from the model~(\fig{fig:results:navigation}B). Models without maps (left) are never shown any intersection data during training and cannot make any turns. When testing navigation-trained models but without any route (right), the model has no way to figure out which direction to go and randomly picks a direction. However, as it enters the intersection it quickly alternates between committing to a left or right execution, and ultimately completes a wide, non-stable turn.

We also quantitatively measure the performance at different intersections in the test track. As in \fig{fig:results:intersections}, vehicle steering onto or away from smaller side streets (\eg, intersections 1, 4, 6, 8) is harder than larger side streets (2, 3, 7). Intersection 5 is particularly interesting as the model had difficulties completing the sharp ($<$ 90 degree) turn angle. We observe that our system performs slightly better on right turns over left and hypothesize that these turns are somewhat easier as the car can simply follow the right boundary of the road to complete right turns, but needs to perceive and attend to a more distant opposite boundary during a left turn. This observation is also consistent with previous literature~\cite{shu2020autonomous}.

\input{figText/results/efficiency}

\subsection{Model Efficiency and Speedup}

As in \fig{fig:efficiency}, our optimized sparse convolution kernel achieves a \textbf{3.6$\times$} acceleration, and our model design further reduces the latency by another \textbf{2.6$\times$}. As a result, our \model runs at 47 FPS on NVIDIA GeForce GTX 1080Ti and 11 FPS on NVIDIA Jetson AGX Xavier. In \tab{tab:efficiency}, we compare \model with the previously fastest 3D model (PointNet). With \textbf{12$\times$} computation reduction, our model is still able to drive the vehicle smoothly through the test track while PointNet crashes the vehicle almost instantly. This is mainly because PointNet relies only on the final pooling layer to aggregate the feature and does not have any convolution, which limits its capability of extracting powerful features.

%% file: figText/results/augmentation.tex
\begin{figure}[!b]
\centering
\vspace{-10pt}
\includegraphics[width=1.0\linewidth]{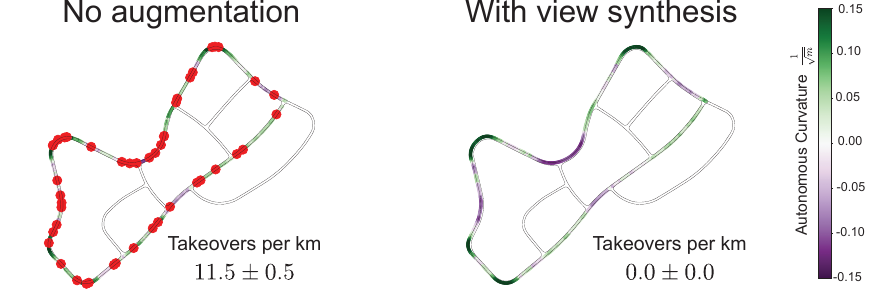}
\caption{\textbf{Rotation augmentation} teaches the model to recover from off-center views. When deployed with the model trained without rotation augmentation, the vehicle crashes frequently.}
\label{fig:results:augmentation}
\end{figure}

%% file: figText/results/recovery.tex
\begin{table}[!t]
\setlength{\tabcolsep}{8pt}
\small\centering
\begin{tabular}{lccc}
\toprule
Model &    CW &   CCW &  Average \\
\midrule
PointNet      & 0.12 $\pm$ 0.17 & 0.00 $\pm$ 0.00 & 0.06 $\pm$ 0.08 \\
Deterministic & 0.80 $\pm$ 0.13 & 0.92 $\pm$ 0.10 & 0.86 $\pm$ 0.08 \\
Hybrid        & 0.80 $\pm$ 0.13 & 0.90 $\pm$ 0.11 & 0.85 $\pm$ 0.08 \\
Uniform       & 0.84 $\pm$ 0.16 & 0.94 $\pm$ 0.10 & 0.89 $\pm$ 0.07 \\
Evidence      & 0.86 $\pm$ 0.10 & 0.92 $\pm$ 0.10 & 0.89 $\pm$ 0.07 \\
\bottomrule
\end{tabular}
\caption{\textbf{Performance of recovering from near-crash positions}. Here, CW denotes the success rate of recovering from clockwise rotation, and CCW denotes counter-clockwise.}
\label{tab:recovery}
\end{table}

%% file: figText/results/intersection.tex
\begin{figure}[!t]
\centering
\includegraphics[width=\linewidth]{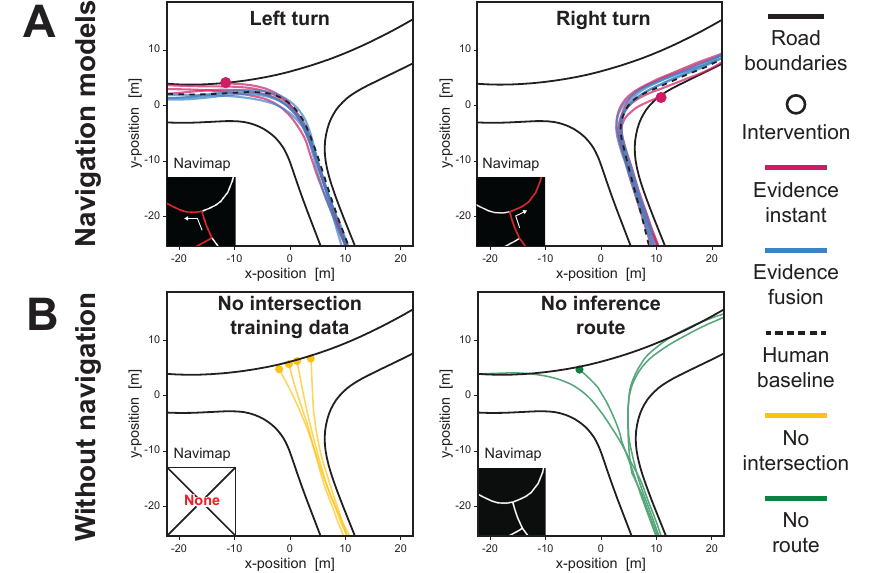}
\caption{\textbf{Evaluation of navigation performance.}  (\textbf{A}) Performance of models with navigation enabled. Evidence fusion models achieve optimal performance on par with human. (\textbf{B}) Performance of models evaluated without navigation. Note that different curves are from different real-world trials.}
\vspace{-10pt}
\label{fig:results:navigation}
\end{figure}

%% file: figText/results/intersections.tex
\begin{figure}[!b]
\centering
\vspace{-10pt}
\includegraphics[width=1.0\linewidth]{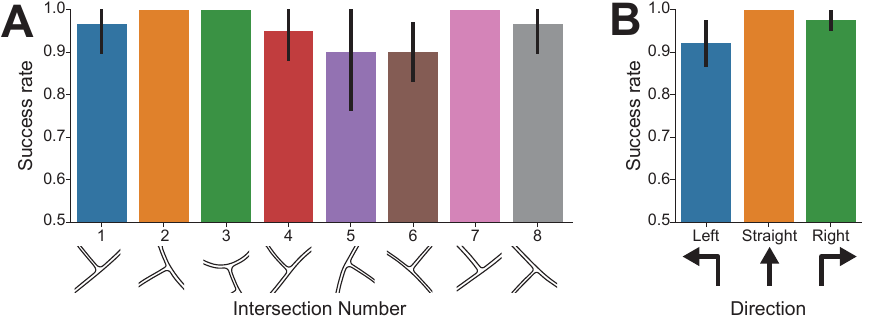}
\caption{\textbf{Success rates at individual intersections.} (\textbf{A}). Success rates for each intersection on the test track. (\textbf{B}) Success rates for different maneuver directions.}
\vspace{-15pt}
\label{fig:results:intersections}
\end{figure}

%% file: figText/results/efficiency.tex
\begin{figure}[!t]
\centering
\includegraphics[width=1.0\linewidth]{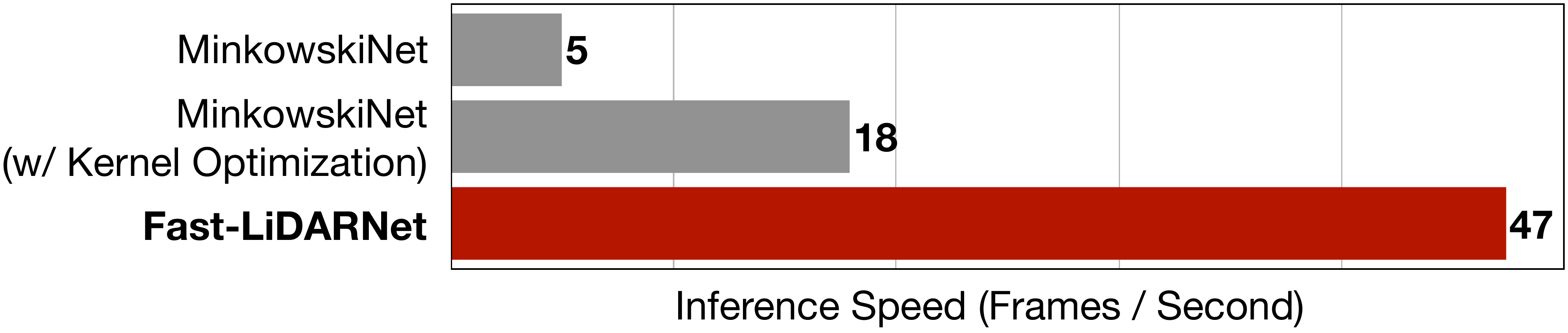}
\caption{\textbf{Sparse convolution kernel optimization and our model design} significantly speed up the model inference.}
\label{fig:efficiency}
\end{figure}

\begin{table}[!t]
\setlength{\tabcolsep}{7pt}
\small\centering
\begin{tabular}{ccrl}
    \toprule
     & PointNet~\cite{qi2017pointnet} & \multicolumn{2}{c}{\model (Ours)} \\
    \midrule
    \#Params & 0.87M & 0.48M & (\textbf{1.6$\times$} compression)  \\
    \#MACs & 7.88G & 0.64G & (\textbf{12$\times$} reduction) \\
    \midrule
    Road Test & Failed & \multicolumn{2}{c}{Succeed} \\
    \bottomrule
\end{tabular}
\caption{\model is parameter-efficient, computation-efficient, and effectively avoids crashing during the road test.}
\vspace{-15pt}
\label{tab:efficiency}
\end{table}

%% file: text/conclusion.tex
\section{Conclusion}
\label{sect:conclusion}

In this paper, we present an efficient and robust LiDAR-based end-to-end navigation framework. We optimize the sparse convolution kernel and redesign the 3D neural architecture to achieve fast LiDAR processing. We also propose the novel \fusion that fuses predictions from multiple frames together while taking their uncertainties into consideration. Our framework has been evaluated on a full-scale autonomous vehicle and demonstrates lane-stable as well as navigation capabilities. Our proposed fusion algorithm significantly improves robustness and reduces the number of takeovers in the presence of out-of-distribution events (\eg sensor failures).